# Wavelet Based QRS Complex Detection of ECG Signal

## Sayantan Mukhopadhyay[1], Shouvik Biswas[2], Anamitra Bardhan Roy[3], Nilanjan Dey[4]

[1,2,3]Department of CSE , JIS College of Engineering, Kalyani,WestBengal,India
[4] Department of IT, JIS College of Engineering, Kalyani, WestBengal, India

**ABSTRACT**

**The Electrocardiogram (ECG) is a sensitive diagnostic tool that is used to detect various cardio-vascular diseases by measuring and recording the electrical activity of the heart in exquisite detail. A wide range of heart condition is determined by thorough examination of the features of the ECG report. Automatic extraction of time plane features is important for identification of vital cardiac diseases. This paper presents a multi-resolution wavelet transform based system for detection 'P', 'Q', 'R', 'S', 'T' peaks complex from original ECG signal. 'R-R' time lapse is an important minutia of the ECG signal that corresponds to the heartbeat of the concerned person. Abrupt increase in height of the 'R' wave or changes in the measurement of the 'R-R' denote various anomalies of human heart. Similarly 'P-P', 'Q-Q', 'S-S', 'T-T' also corresponds to different anomalies of heart and their peak amplitude also envisages other cardiac diseases. In this proposed method the 'P Q R S T'-peaks are marked and stored over the entire signal and the time interval between two consecutive 'R'-peaks and other peaks interval are measured to detect anomalies in behavior of heart, if any. The peaks are achieved by the composition of Daubeheissub-bands wavelet of original ECG signal. The accuracy of 'P Q R S T' complex detection and interval measurement is achieved up to 100% with high exactitude by processing and thresholding the original ECG signal.**

## I.    INTRODUCTION

  Myocardial infarction (MI) or acute myocardial infarction (AMI) [1], commonly known as a heart attack is the most frequently occurring disease worldwide. This is also the major cause of the large number of premature deaths globally. This happens generally due to the occlusion of the coronary artery or pulmonary veins, which is triggered by natural, social, genetic or human behavior. If left untreated for a sufficient period, can cause infarction of myocardium (heart muscle tissue) and lead to fatal consequences.

Electrocardiogram is a classical diagnosis to detect heart malfunctioning and heart muscle damage. This can even prognosticate heart attack in severe case and is a unique tool for diagnosing several cardiac diseases like bradycardia, tachycardia, cardiac arrhythmias, conduction abnormalities, ventricular hypertrophy, myocardial infection and other disease states [2].

An ECG signal is composed of successive repetition of 'PQRST' in monotony. In the beginning, a crust is generated from the linear signal to form the 'P' wave. The declining linear wave soon gets a downward deflection labeled as 'Q' wave. A sudden upright deflection can be observed just beyond the Q wave to form a high cone that is, the 'R' wave. On its decline a slight downward defection is the 'S' wave. A noticeable hinge after the 'S' wave is known as 'T' wave that marks the end of a segment of the ECG signal. [3]

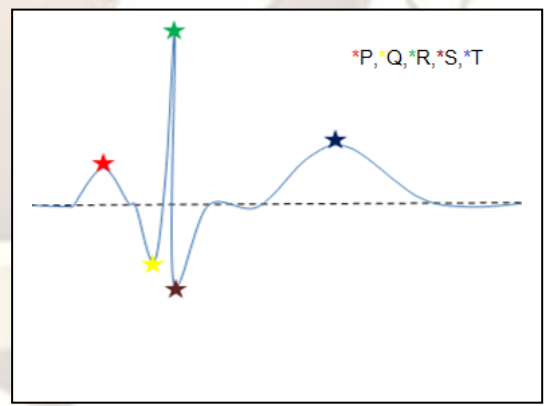

Fig 1. P-QRS-T

In this paper 'R-R' interval is emphasized for analyzing the heart condition of the concerned person based on the study of the ECG signal. For this, the 'QRS' wave is located in the ECG signal from which the 'R' wave is manifested. Considering the time axis along the ECG signal the time lapse between two consecutive R waves is measured [4]. This algorithm detects the R-R interval automatically by decomposing Daubechies Wavelet (db6) as mother wavelet generated from the ECG signal.

## II.    METHODOLOGY

### A.  DISCRETE WAVELET TRANSFORMATION

The wavelet transform describes a multi-resolution decomposition process in terms of expansion of a signal onto a set of wavelet basis functions. Discrete Wavelet Transformation has its own excellent space frequency localization property. Application of DWT in 1D signal corresponds to 1D filter in each dimension. The input Daubechies Wavelet as mother wavelet is divided into 8 non-overlapping multi-resolution sub-bands by the filters, namely db1, db2, db3up to db8, where db is acronym for Daubechies. The sub-band is processed further to obtain the next coarser scale of wavelet coefficients, until some final scale "N" is reached. When a signal is decomposed into 8 levels, the db6 sub-band best reflects the original signal, since according to the wavelet theory, the approximation signal at level n is the aggregation of the approximation at level n-1 plus the detail at level n-1[5].

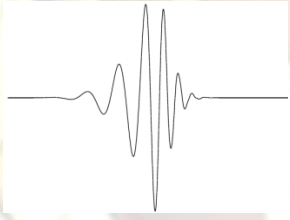

Figure 2. Daubechies Wavelet

## III.    PROPOSED METHOD

Step1: ECG signal is read and the length is calculated.

Step2: The signal is decomposed using db6 wavelet.

Step 3: $3^{rd}$, $4^{th}$ and $5^{th}$ detail coefficients are selected, as most energy of the QRS complex is concentrated in these coefficients.

Step 4: The wave is reconstructed using detail coefficient 3, 4, 5.(D1=d3+d4+d5)

Step 5: A function d4*(d3+d5) /$2^n$ is defined to reduce the oscillatory nature of the signal where d3, d4, d5 are the 3th, 4th ,5th detail coefficients and n is the level of decomposition.

Step6: Derivate up to level 5 is made using the transfer function

$$H(z) = \left(\frac{T}{8}\right)(-z^{-2} - 2z^{-1} + 2z^1 + z^2)$$

Step7: The differential equation:

y(nT)=(T/8)(-x(nT-2T)-2x(nT-T)+2x(nT+T)+x(nT+2t)

is applied to the signal, by using the transfer function and taking the amplitude response as

$$|H(wT)| = \left(\frac{T}{4}\right)[sin(2\omega T) + 2sin(\omega T)]$$

Step8: The signal is squared point by point using the equation

y(nT)=(T/8)(-x(nT-2T)-2x(nT-T)+2x(nT+T)+x(nT+2t))

to emphasize R wave from the ECG signal.

Step9: A moving window is integrated using the equation

Y=(1/N)*[x(nT-(N-1)T)+x(nT-(N-2)T)+…+x(nT)]

to obtain the waveform feature information.

Step10: The threshold value is calculated corresponding to the product of max and mean of the signal to locate the end points of the moving window.

Step11: The PQRST peaks are located based on the amplitude of the signal within each moving window.

Step12: The time intervals are calculated considering the positions of two consecutive same labeled peaks and stored

Step13: Diagnosis of various cardiac diseases is done by comparing ground truth conditions with the data.

## IV. EXPLANATION OF THE PROPOSED METHOD

ECG signal decomposed using db6 wavelet. The 3rd , 4th and 5th  detail coefficients are summed up for reconstruction of the wave, as most energy of the QRS complex is concentrated in these coefficients.. A function D2=d4*(d3+d5) /$2^n$ is defined to reduce the oscillatory nature of the signal where  d3, d4, d5 are the detail coefficients and n is the level of decomposition[6]. Derivation of the ECG wave is done to obtain the QRS wave segment information. Derivate up to level 5 is made using the transfer function

$$H(z) = \left(\frac{T}{8}\right)(-z^{-2} - 2z^{-1} + 2z^1 + z^2)$$

...... (1)

The differential equation:

y(nT)=(T/8)(-x(nT-2T)-2x(nT-T)+2x(nT+T)+x(nT+2t)

..… (2)

is applied to the signal, by using the transfer function and taking the amplitude response as

$$|H(wT)| = \left(\frac{T}{4}\right)[sin(2\omega T) + 2sin(\omega T)]$$

….(3)

This gives close approximation of ideal derivative over this range, with negligible delay.

The important mathematical concept of Squaring is used to nullify the low amplitude waves like and around P and T waves.

The signal is squared point-by-point using the equation

y(nT)=(T/8)(-x(nT-2T)-2x(nT-T)+2x(nT+T)+x(nT+2t)

… (4)

This gives magnified resultant of the emphasized R wave over the entire range of the ECG signal

A moving window is integrated using the equation

Y=(1/N)*[x(nT-(N-1)T)+x(nT-(N-2)T)+…+x(nT)]
…… (5)

to obtain the waveform feature information. The sample size i.e. N varies for different cases and the wideness range of QRS complex also lies over a vast measurement. If the window is wide enough then the QRS complex gets merge with the T complex. Whereas, in case of narrow window several QRS peaks is generated to increase the complexity of analysis. Hence, the concept of moving window is induced in the algorithm to accustomed different samples for analysis.

## V. RESULT AND DISCUSION

The sample ECG signal analyzed in this method as test cases are rendered by Suraha Nursing Home,Kolkata.

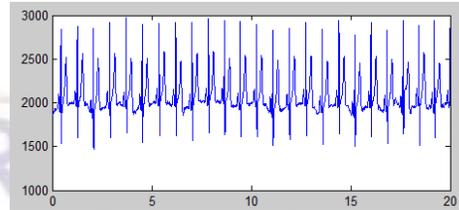

Fig 3. Original ECG Signal

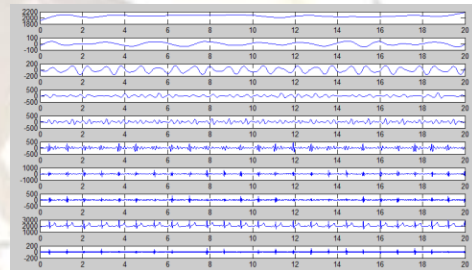

Fig 4.Daubehies wavelet (db6) Decomposed ECG Signal

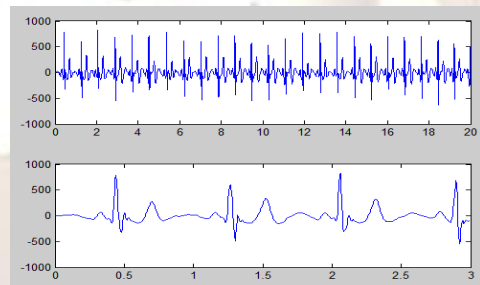

Fig 5.Plot of D1 (d3+d4+d5)

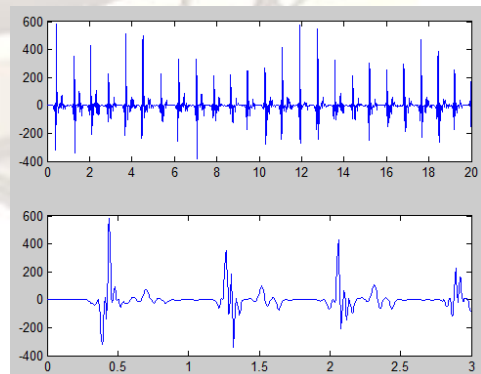

Fig 6.Plot of D2 (d4*(d3+d5) /$2^n$)

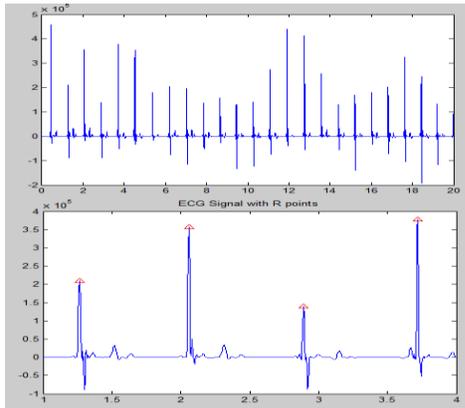

Fig 7.Plot of ECG signal with R Peaks

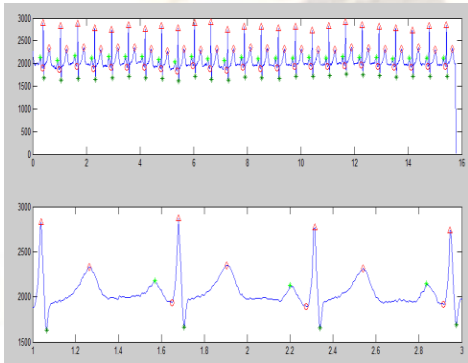

Fig 8.Plot of ECG signal with P-QRS-T

The time interval between R-R is 0.6450 sec.

Table1: Disease prediction based on Intervals of Peaks

| INTERVALS | Normal Sinus Rhythm (NRS) in sec | <NRS Diseases | >NRS Diseases |
|---|---|---|---|
| P-R | 0.12-0.2 | Reduced FMD; | Blockage of AV node; atherosclerotic disease |
| Q-R-S | 0.09 | Hypercalemia | - |
| Q-T | 0.35-0.44 | - | - |
| S-T | 0.05-0.15 | - | - |
| P-P | 0.11 | - | - |
| R-R | 0.80-0.85 | Tachycardia (Fast heart) | Bradycardia (Slow heart) |

Table2: Disease Prediction Based on Amplitude of Peaks

| AMPLITUDE | Normal Sinus Rhythm( NRS) in mV | <NRS | >NRS |
|---|---|---|---|
| P | 0.25 | Dextrocardia (inverted P wave) | - |
| Q | 25% of R wave | - | - |
| R | 1.60 | - | - |
| S | - | - | - |
| T | 0.1-0.5 | Myocardial ischemia (inverted T wave) | Hyperkalemia (Tall T wave & absence of P wave) |

Beyond this numerical aspect Sinoatrial block is traced from complete drop out of a cardiac cycle. In addition, Irregular ECG can prophesy to the extent of sudden cardiac death.

## VI. CONCLUTION

Since the application of wavelet transformation in electrocardiology is relatively new field of research, many methodological aspects (choice of the mother wavelet, values of the scale parameters) of the wavelet technique will require further investigations in order to improve the clinical usefulness of this original signal processing technique. Imperative clinical usefulness is drawn from the innovative application of wavelet transformation in electrocardiology. Since this field of research is new and fostering, the wavelet technique will require auxiliary investigations in many methodological aspects such as selecting choice of mother wavelet and values of scale parameter to improve the clinical utility. This biomedical signal processing resulted in detecting many cardiovascular diseases based on diagnosing the ECG signal. The measurement of amplitude and intervals achieved in the proposed method is hence compared to the ground truth by medical expertise to detect a set of cardio vascular diseases in human.